\documentclass[11pt, a4paper]{article}

\usepackage[utf8]{inputenc}
\usepackage[T1]{fontenc}
\usepackage{geometry}
\usepackage{amsmath, amssymb, amsfonts}
\usepackage{graphicx}
\usepackage{booktabs} 
\usepackage{multirow}
\usepackage{xcolor}
\usepackage{xurl} 
\usepackage{hyperref}
\usepackage{cite} 
\usepackage{caption}
\usepackage{subcaption} 
\usepackage{amsthm}

\hypersetup{
    colorlinks=true,
    linkcolor=blue,
    filecolor=magenta,      
    urlcolor=cyan,
    citecolor=blue,
}

\usepackage{tikz}
\usetikzlibrary{shapes.geometric, arrows.meta, positioning, fit, calc, backgrounds}

\usepackage{fancyhdr}
\makeatletter
\newcommand\blfootnote[1]{%
  \begingroup
  \renewcommand\thefootnote{}\footnote{#1}%
  \addtocounter{footnote}{-1}%
  \endgroup
}
\makeatother

\makeatletter
\renewcommand{\maketitle}{
  \bgroup\setlength{\parindent}{0pt}
  \begin{flushleft}
    \vspace*{-1.5cm} 
    \includegraphics[height=0.96cm]{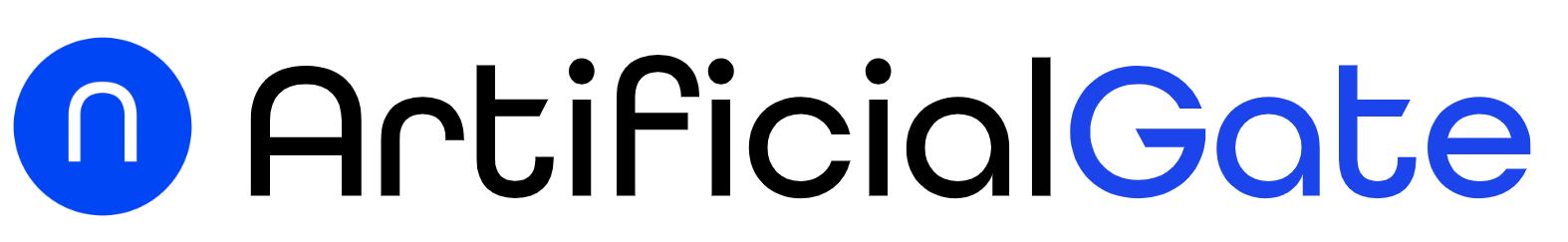} 
    \vspace{0.3cm} \\
    \rule{\textwidth}{1.2pt} \\ 
    \vspace{0.8cm}
    {\LARGE \sffamily \textbf{\@title} \par} 
    \vspace{0.6cm}
    {\large \@author \par}
  \end{flushleft}
  \egroup
  \vspace{0.8cm}
  \thispagestyle{empty} 
}
\makeatother

\title{Quantifying Prior Dominance in RAG Systems}

\author{
  \textbf{Barak Or} \\
  \small ArtificialGate Ltd. \\
  \texttt{barak@artificialgate.ai}
}
\date{} 

\begin{document}

\maketitle

\blfootnote{This research was independently conducted and funded by ArtificialGate Ltd. The author serves as Academic Director at the Google \& Reichman Tech School, and as an External Lecturer at Technion -- Israel Institute of Technology and Reichman University.}

\begin{abstract}
Retrieval-Augmented Generation (RAG) grounds Large Language Models in external knowledge, yet current evaluations rely on discrete heuristics that suffer from ''epistemic blindness'' - failing to distinguish genuine contextual information extraction from parametric memory recall. To address this, we introduce the Normalized Context Utilization (NCU) metric, leveraging continuous token log-probabilities across zero-shot, oracle, and adversarial conditions to strictly quantify contextual information gain. Evaluating architectures ranging from 1.5B to 72B parameters alongside a proprietary commercial API reveals that for strict factual extraction (without Chain-of-Thought reasoning), traditional scaling laws exhibit extreme diminishing returns: highly efficient Small Language Models (SLMs) match or outperform high-capacity architectures. Furthermore, we demonstrate that ``Prior Dominance'' correlates with model scale and proprietary alignments. The evaluated commercial API not only overrode explicit external evidence in nearly half of adversarial conflicts, but also frequently suffered from systemic confidence collapse (\textit{Negative Transfer}) when its parametric priors were contradicted. Our findings highlight the structural epistemic advantage and superior contextual adherence of SLMs in strict extraction workflows.
\end{abstract}

\section{Introduction}
\label{sec:intro}

As Large Language Models (LLMs) achieve broad general-purpose reasoning capabilities, integrating dynamic, out-of-distribution knowledge remains a fundamental challenge. The prevailing paradigms for knowledge integration-pre-training foundational models or applying supervised Fine-Tuning (FT)-are bottlenecked by compute constraints, high-latency updates, and catastrophic forgetting. Consequently, Retrieval-Augmented Generation (RAG) \cite{lewis2020retrieval, borgeaud2022improving, guu2020realm, karpukhin2020dense} has emerged as a standard mechanism for decoupling knowledge from model weights, aiming to shift the computational burden from parametric memory recall to active contextual processing.

The trajectory of LLM development is largely guided by established scaling laws \cite{kaplan2020scaling, hoffmann2022training, xiong2024scaling, wang2024scaling}, which posit that larger parameter counts predictably yield better downstream performance. Applying this assumption directly to RAG architectures, however, exposes a methodological limitation in current evaluation frameworks. Standard open-domain Question Answering (QA) benchmarks rely heavily on discrete text-matching heuristics, such as Exact Match (EM) or F1 scores \cite{papineni2002bleu, lin2004rouge, min2023factscore}. These binary metrics suffer from an evaluation blind spot: when a model correctly answers a query given a retrieved context, standard heuristics cannot ascertain whether the model genuinely extracted the target information from the text, or merely recited the answer from its pre-trained weights. 

This ambiguity obfuscates true contextual utilization. The apparent superiority of massively scaled models in RAG benchmarks may stem from their extensive prior knowledge rather than superior contextual processing. Furthermore, binary metrics fail to capture continuous fluctuations in a model's predictive confidence when encountering adversarial contexts, knowledge conflicts \cite{chen2023rich, longpre2021entity, mallen2023when}, or semantic noise. 

To rigorously disentangle parametric memory from active context extraction, we propose a shift from discrete text matching to continuous probability spaces. In this paper, we introduce the \textbf{Normalized Context Utilization (NCU)} metric. Unlike traditional heuristics, NCU leverages continuous length-normalized log-probabilities across zero-shot, oracle, and adversarial conditions. By isolating the generator component with provided oracle contexts-effectively evaluating Context-Augmented Generation (CAG)-we can rigorously assess the fractional reduction in logarithmic uncertainty when a model is exposed to contextual evidence. Our study specifically focuses on strict factual extraction without Chain-of-Thought (CoT) reasoning paradigms.

\begin{figure}[htb!]
    \centering
    \resizebox{\textwidth}{!}{
    \begin{tikzpicture}[
        node distance=1.2cm and 1.5cm,
        font=\sffamily\small, 
        base/.style={align=center, inner sep=2.5mm, outer sep=0pt, rounded corners=4pt}, 
        model/.style={base, fill=blue!50, text=white, minimum width=3.2cm, minimum height=1cm, font=\sffamily\small\bfseries},
        data_in/.style={base, fill=cyan!15, text=black, minimum width=2.8cm}, 
        data_out/.style={base, fill=purple!15, text=black, minimum width=3.2cm},
        metricBad/.style={base, fill=red!20, text=black, minimum width=3.2cm, font=\sffamily\small\bfseries},
        metricGood/.style={base, fill=teal!30, text=black, minimum width=4cm, font=\sffamily\small\bfseries},
        arrow/.style={-{Stealth[scale=1.1]}, line width=1pt, draw=black!70},
        dashed_arrow/.style={-{Stealth[scale=1.2]}, line width=1.5pt, dashed, draw=black!50}
    ]

    \node (q1) [data_in] {Query + Context};
    \node (llm1) [model, below=of q1] {LLM (Standard)};
    \node (out1) [data_out, below=of llm1] {Text Output};
    \node (eval1) [metricBad, below=of out1] {Discrete Match (EM)};
    
    \draw [arrow] (q1.south) -- (llm1.north);
    \draw [arrow] (llm1.south) -- (out1.north);
    \draw [arrow] (out1.south) -- (eval1.north);

    \node (llm2) [model, fill=teal!60, minimum width=5.5cm, right=6.8cm of llm1] {LLM Evaluation (Any Scale)};
    
    \node (q_zero) [data_in, above=1cm of llm2, xshift=-2.1cm] {Zero-Shot Query};
    \node (q_oracle) [data_in, above=1cm of llm2, xshift=2.1cm] {Oracle Context};
    
    \node (log_prior) [data_out, below=1cm of llm2, xshift=-2.1cm] {Prior:\\ $\log P_\theta(Y|Q)$};
    \node (log_post) [data_out, below=1cm of llm2, xshift=2.1cm] {Posterior:\\ $\log P_\theta(Y|Q,C)$};
    
    \node (ncu) [metricGood, minimum width=5.5cm] at (llm2 |- eval1) {Continuous NCU Metric};

    \draw [arrow] (q_zero.south) -- (llm2.north -| q_zero.south);
    \draw [arrow] (q_oracle.south) -- (llm2.north -| q_oracle.south);
    
    \draw [arrow] (llm2.south -| log_prior.north) -- (log_prior.north);
    \draw [arrow] (llm2.south -| log_post.north) -- (log_post.north);
    
    \draw [arrow] (log_prior.south) -- (ncu.north -| log_prior.south);
    \draw [arrow] (log_post.south) -- (ncu.north -| log_post.south);

    \begin{scope}[on background layer]
        \node (box1) [draw=red!50, line width=1.5pt, rounded corners=6pt, inner sep=6mm, fit=(q1) (eval1)] {};
        \node [above=2mm, font=\sffamily\bfseries\color{red!70!black}] at (box1.north) {Standard Evaluation};

        \node (box2) [draw=teal!60, line width=1.5pt, rounded corners=6pt, inner sep=6mm, fit=(q_zero) (q_oracle) (ncu) (log_prior) (log_post)] {};
        \node [above=2mm, font=\sffamily\bfseries\color{teal!70!black}] at (box2.north) {NCU Probability Framework};
    \end{scope}

    \draw [dashed_arrow] (eval1.east) -- node[above, midway, align=center, font=\sffamily\bfseries, text=red!80!black, fill=white, inner sep=2pt, yshift=1.5mm] {Evaluation\\Blind Spot} (ncu.west);

    \end{tikzpicture}
    }
    \caption{\textbf{Conceptual Architecture: Standard Evaluation vs. NCU Framework.} Traditional metrics (left) rely on binary text matching, failing to distinguish parametric recall from reading comprehension. The NCU framework (right) utilizes continuous log-probabilities across zero-shot and oracle conditions to isolate the informational gain of the RAG context.}
    \label{fig:concept_diagram}
\end{figure}
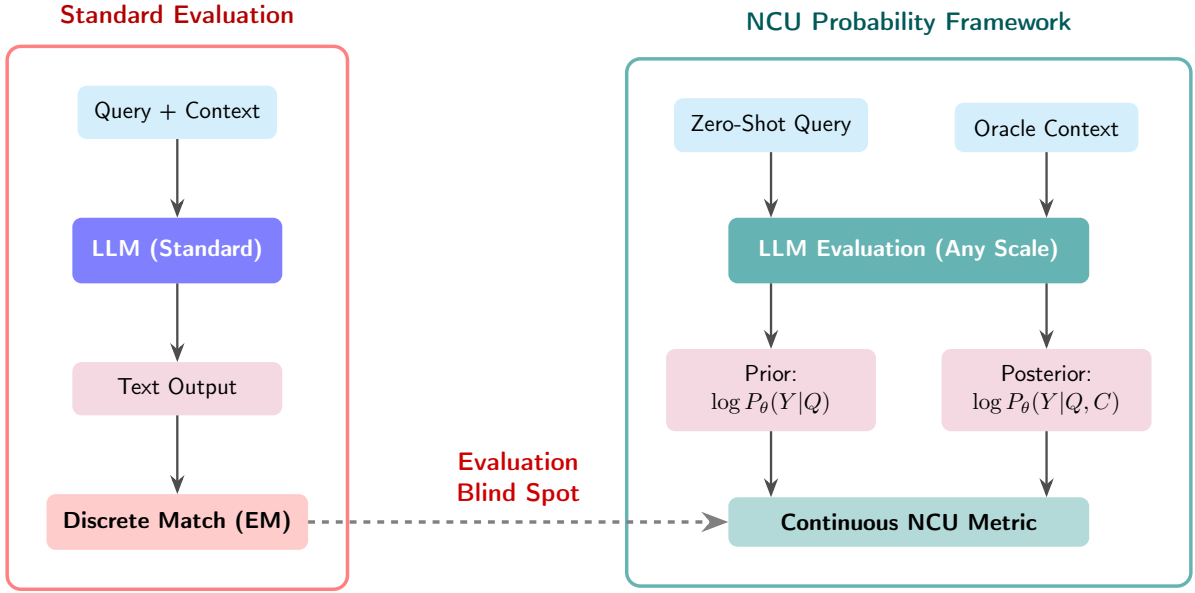

To empirically explore this framework, we conducted bounded inferences across three distinct QA domains. By evaluating architectures ranging from 1.5B open-weight Small Language Models (SLMs) to massive models (72B) and a proprietary commercial API against systematically perturbed contexts, we mapped the empirical boundaries of context utilization. 

\begin{figure}[htb!]
    \centering
    \resizebox{\textwidth}{!}{
    \begin{tikzpicture}[
        font=\sffamily\small,
        box/.style={align=center, inner sep=3mm, rounded corners=6pt, line width=1pt},
        input/.style={box, fill=cyan!15, draw=cyan!50, minimum width=3.5cm},
        output/.style={box, fill=purple!15, draw=purple!50, minimum width=4.5cm},
        model/.style={box, draw=black!70, text=white, font=\sffamily\small\bfseries},
        bubble/.style={box, draw=black!40, fill=white, inner sep=2mm, font=\sffamily\scriptsize, align=center, rounded corners=2pt},
        arrow/.style={-{Stealth[scale=1.1]}, line width=1.2pt, draw=black!70},
        red_arrow/.style={-{Stealth[scale=1.1]}, line width=1.5pt, draw=red!70},
        teal_arrow/.style={-{Stealth[scale=1.1]}, line width=1.5pt, draw=teal!70}
    ]

    \node (raw_l) [input] at (0, 6) {\textbf{Adversarial Context ($C_{adv}$)} \\ Fact: $Y_{adv}$};
    
    \node (giant) [model, fill=blue!50, minimum width=6.8cm, minimum height=4.5cm] at (0, 0) {};
    \node (giant_label) [align=center] at (0, 1.5) {High-Capacity Model \\ (e.g., Proprietary APIs)};
    \node (boulder) [draw=black!70, fill=gray!30, inner sep=2mm, minimum width=5.2cm, minimum height=1.2cm, align=center, text=black, font=\sffamily\small\bfseries] at (0, -0.8) {Parametric Weights};
    
    \node (instr_l) [bubble, rounded corners=4pt] at (-6.0, 0) {\textbf{Instruction} \\ Synthesize strictly \\ from context};
    \node (out_l) [output] at (0, -5.5) {Output: $Y_{true}$ \\ \textbf{Prior Dominance}};
    
    \node (loud) [bubble, fill=red!5, draw=red!50] at (5.8, -0.8) {Weight Distribution: \\ $\mathbf{P(Y_{true}) \gg P(Y_{adv}|C_{adv})}$};

    \draw [arrow, dashed] (raw_l.south) -- node[circle, solid, draw=red!70, fill=white, inner sep=2pt, font=\sffamily\small\bfseries\color{red!70}] {Reject} (giant.north);
    \draw [arrow] (instr_l.east) -- (giant.west);
    \draw [red_arrow, <->] (loud.west) -- (boulder.east);
    \draw [red_arrow] (boulder.south) -- (out_l.north);

    \node (raw_r) [input] at (17, 6) {\textbf{Adversarial Context ($C_{adv}$)} \\ Fact: $Y_{adv}$};
    
    \node (slm) [model, fill=teal!60, minimum width=4.5cm, minimum height=3.5cm] at (17, -0.5) {};
    \node (slm_label) [align=center] at (17, 0.5) {Low-Capacity Model \\ (SLM 1.5B - 7B)};
    \node (slate) [fill=white, draw=teal!60, dotted, minimum width=3cm, minimum height=0.8cm, align=center, text=teal!70, font=\sffamily\scriptsize\bfseries] at (17, -1.2) {Weak-Prior Processor};
    
    \node (instr_r) [bubble, rounded corners=4pt] at (12.0, -0.5) {\textbf{Instruction} \\ Synthesize strictly \\ from context};
    \node (out_r) [output] at (17, -5.5) {Output: $Y_{adv}$ \\ \textbf{Context Adherence}};
    
    \node (quiet) [bubble, fill=teal!5, draw=teal!50] at (22.5, -1.2) {Weight Distribution: \\ $\mathbf{P(Y_{adv}|C_{adv}) \gg P(Y_{true})}$};

    \draw [teal_arrow] (raw_r.south) -- (slm.north);
    \draw [arrow] (instr_r.east) -- (slm.west);
    \draw [teal_arrow, dotted] (quiet.west) -- (slate.east);
    \draw [teal_arrow] (slm.south) -- (out_r.north);

    \end{tikzpicture}
    }
    \caption{\textbf{Prior Dominance vs. Context Adherence.} A conceptual visualization of knowledge conflicts. Left: High-capacity models may override contradictory context due to heavily weighted parametric priors ($P_{param} \gg P_{ctx}$). Right: Low-capacity SLMs generally exhibit higher context adherence, processing external evidence with fewer parametric constraints.}
    \label{fig:giant_vs_slm_teaser}
\end{figure}
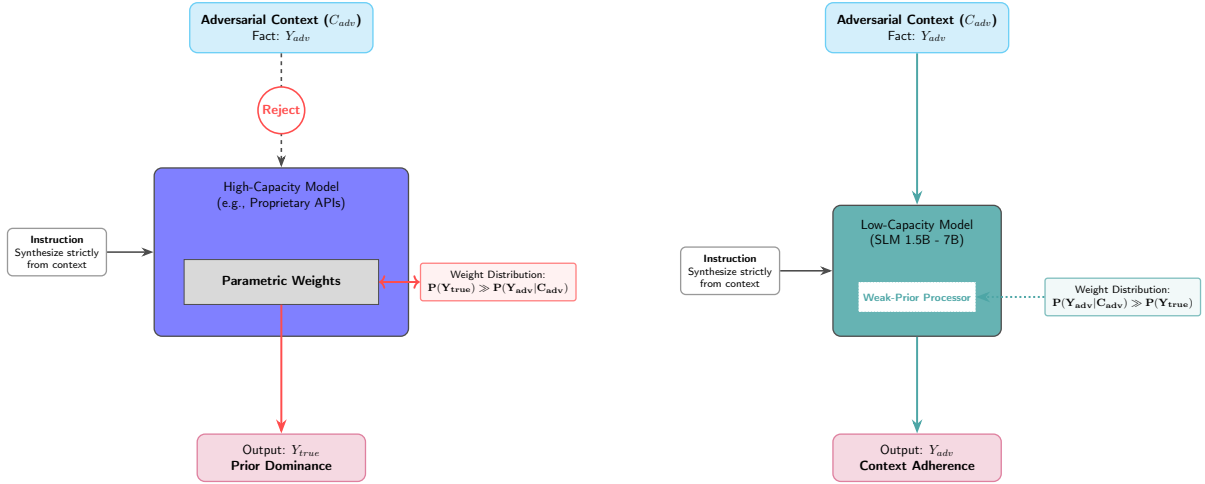

Specifically, this study addresses the following core Research Questions (RQs), conceptually illustrated in \textbf{Figure \ref{fig:giant_vs_slm_teaser}}:
\begin{itemize}
    \item \textbf{RQ1 (Extraction Scaling):} Does increasing parameter count intrinsically improve contextual extraction, or do strict extraction tasks exhibit diminishing returns under standard scaling paradigms?
    \item \textbf{RQ2 (Prior Dominance):} How do parametric scale and proprietary alignments influence a model's propensity to override explicitly provided, yet contradictory, external evidence?
    \item \textbf{RQ3 (Negative Transfer):} Under what conditions does semantic conflict cause high-capacity models to suffer from negative transfer (i.e., information loss relative to their zero-shot baseline)?
\end{itemize}

Through our empirical analysis, \textbf{we present the following core contributions:}
\begin{enumerate}
    \item \textbf{Continuous Evaluation Metric:} We define and validate the NCU metric, providing a probability-driven framework that isolates contextual processing from parametric memory, normalizing for vocabulary and tokenization disparities across models.
    \item \textbf{Diminishing Returns in Extraction Scaling:} We empirically demonstrate severe diminishing returns for parameter scale in factual grounding tasks. SLMs (1.5B--7B) achieved statistical parity in context utilization compared to a 72B parameter model, while offering up to a $70\times$ reduction in inference latency relative to a commercial API.
    \item \textbf{Quantification of Prior Dominance and Negative Transfer:} We identify a measurable vulnerability in high-capacity models: when presented with explicit external conflicts, they frequently override context (\textit{Prior Dominance}) and are susceptible to systemic confidence collapse (\textit{Negative Transfer}), whereas SLMs maintain higher epistemic stability.
\end{enumerate}

The remainder of this paper is structured as follows. Section \ref{sec:related_work} reviews the literature on RAG evaluation and scaling laws. Section \ref{sec:theory} frames the theoretical metric. Section \ref{sec:experimental_setup} details our methodology. Section \ref{sec:results} presents our empirical findings. Section \ref{sec:discussion} discusses the operational implications. Section \ref{sec:limitations} acknowledges limitations, and Section \ref{sec:conclusion} concludes.

\section{Related Work}
\label{sec:related_work}

The efficacy of RAG is fundamentally tied to a model's ability to prioritize external evidence over internal priors. This section contextualizes our NCU framework within the literature regarding RAG evaluation, scaling laws, and knowledge conflict resolution.

\subsection{The Evolution of RAG Evaluation}
Early RAG architectures were predominantly evaluated using discrete lexical metrics such as EM and F1-score \cite{lewis2020retrieval}. As LLMs evolved, evaluation paradigms shifted toward LLM-as-a-judge frameworks \cite{es2023ragas, saadfalcon2023ares, zheng2023judging}. However, recent findings have highlighted the limitations of these discrete assessments, noting that binary heuristics suffer from "epistemic blindness" \cite{gu2026redefining}. Consequently, evaluation is pivoting toward Information Theoretic benchmarks and token-level log-probability tracking \cite{xiao2026revisiting, sun2025rageval}. Our continuous NCU metric builds upon this foundation by formalizing the fractional reduction in predictive uncertainty. While "LLM-as-a-judge" frameworks like RAGAS and ARES offer progress beyond exact matching, they often rely on the reasoning of another black-box model, which can introduce its own biases and lack of transparency. In contrast, our NCU metric leverages direct token-level log-probabilities, providing a more objective, reproducible, and fine-grained measure of how the model's internal confidence shifts when exposed to external context.

\subsection{Scaling Laws: Pre-training vs. Inference Extraction}
Standard scaling laws \cite{kaplan2020scaling, hoffmann2022training} generally assume that larger parameter counts improve downstream capabilities. However, recent studies on RAG indicate more complex dynamics. Inference scaling laws for RAG reveal non-linear relationships between generator size and retrieval efficacy \cite{zhou2025inference}. Concurrently, literature suggests a trade-off between parametric memorization and a model's willingness to integrate dynamic retrieval \cite{ablin2026memorize}. 

\subsection{The Resurgence of Small Language Models}
The computational overhead of massive LLMs has catalyzed interest in SLMs \cite{touvron2023llama, gunasekar2023textbooks}. Contemporary pipelines increasingly position SLMs as dynamic output rewriters \cite{vernikos2024slm}, collaborative reasoning routers, or efficient context steering \cite{shi2025collab}. Furthermore, fine-tuning SLMs for structured extraction tasks has empirically shown competitive efficacy compared to zero-shot prompting of massive counterparts \cite{li2025slm, huang2025rag, jiang2024ragstar}. Our study aims to offer a quantitative perspective on this architectural shift, highlighting the potential contextual adherence of SLMs in pure extraction workflows.

\subsection{Knowledge Conflicts and Epistemic Stubbornness}
The intersection of parametric memory and external retrieval often leads to knowledge conflicts \cite{longpre2021entity, chen2023rich, ji2023survey}. While the "Lost in the Middle" phenomenon established that LLMs may ignore context due to spatial constraints \cite{liu2023lost}, recent literature explores epistemic suppression \cite{xie2023adaptive}. Benchmarks and resolution frameworks \cite{zhao2026exploring, diao2026seeing} observe that highly parameterized models can over-trust internal weights, sometimes rejecting contradictory external evidence \cite{alansari2026large, tong2025hallucination, zhou2024rag, vu2023freshllms}. Our research synthesizes these observations by mathematically measuring this vulnerability as \textit{Prior Dominance}, while recognizing the hypothesized role of alignment mechanisms in shaping this behavior.

While existing literature increasingly acknowledges the limitations of discrete metrics and the complex behavior of high-capacity models during knowledge conflicts, a unified, continuous metric that robustly isolates information extraction from parametric memory-while normalizing across differing tokenizers-remains underexplored. This study addresses this gap by proposing the length-normalized NCU framework.

\section{Epistemic Uncertainty and Context Utilization}
\label{sec:theory}

To quantify a model's contextual processing, we transition from discrete heuristics to continuous probability spaces. Building upon Shannon's foundational communication theory \cite{shannon1948mathematical, cover1999elements}, we frame RAG through the lens of information theory. In this paradigm, the external context acts as a secondary information channel designed specifically to resolve the model's predictive uncertainty regarding a user query. This continuous monitoring of internal uncertainty aligns with emerging paradigms that model reasoning as a stochastic inference process susceptible to runtime cognitive drift \cite{or2026kalman}.

Let $Q$ denote a user query, $Y = (y_1, y_2, \dots, y_k)$ the ground-truth target sequence composed of $k$ tokens, and $\theta$ the parameters of an LLM. Because different models employ different tokenization schemes (vocabularies), directly comparing the raw sum of log-probabilities across architectures is mathematically flawed. Therefore, we normalize the joint probability by the token sequence length to derive the average log-probability per token. 

In a zero-shot setting, the model's confidence is governed strictly by its parametric prior. To capture this information content uniformly, we map the length-normalized probabilities to Shannon Entropy, defining the prior epistemic uncertainty ($S_{prior}$) as:
\begin{equation}
    S_{prior} = -\frac{1}{k} \sum_{i=1}^k \log P_\theta(y_i | y_{<i}, Q).
\end{equation}

When a retrieval pipeline injects external context $C$, the model computes a posterior distribution. We similarly define the posterior uncertainty ($S_{post}$) as:
\begin{equation}
    S_{post} = -\frac{1}{k} \sum_{i=1}^k \log P_\theta(y_i | y_{<i}, Q, C).
\end{equation}

The raw Context Utilization Score (CUS) represents the absolute informational gain provided by the RAG pipeline. It is mathematically equivalent to the absolute reduction in predictive uncertainty, $CUS = S_{prior} - S_{post}$, explicitly defined as the difference between the normalized log-probabilities:
\begin{equation}
    CUS = \left( \frac{1}{k} \sum_{i=1}^k \log P_\theta(y_i | y_{<i}, Q, C) \right) - \left( \frac{1}{k} \sum_{i=1}^k \log P_\theta(y_i | y_{<i}, Q) \right)
\end{equation}

To create a comparative and standardized metric across models of varying scales and pre-training distributions, we formalize the \textbf{Normalized Context Utilization (NCU)}. 

The absolute information gain is captured by the $CUS$. Because probability is strictly bounded ($P \le 1$), the minimum possible posterior uncertainty is perfect absolute confidence, where $-\log(1) = 0$. Consequently, the theoretical upper bound for this informational gain is strictly limited by the model's initial ignorance: $\max(CUS) = S_{prior} - 0 = S_{prior}$.

Expressing the measured information gain as a fraction of this maximum theoretical bound yields the normalized resolution of uncertainty. To ensure numerical stability, specifically preventing undefined asymptotic behavior when the model is already perfectly confident in a zero-shot setting ($P_\theta(Y | Q) \to 1$, meaning $S_{prior} \to 0$), we introduce a minute smoothing factor $\epsilon$ (e.g., $10^{-5}$) to the denominator. To rigorously handle boundary conditions, including instances of \textit{negative transfer} where adversarial or noisy context exacerbates uncertainty ($S_{post} > S_{prior}$), we apply strict bounding operators. We formally define the NCU metric as follows:
\begin{equation}
    NCU = \max \left[0, \min \left( 1, \frac{CUS}{S_{prior} + \epsilon} \right) \right].
\end{equation}

This dual-normalization strategy strictly isolates the mechanical efficacy of the RAG integration from the model's pre-existing parametric weights. It ensures an equitable evaluation paradigm regardless of the underlying tokenization vocabulary or the model's initial prior knowledge. We optionally track the unclipped raw ratio ($\frac{CUS}{S_{prior}+\epsilon}$) to highlight asymptotic confidence collapse (Negative Transfer).

The formulated metric exhibits desirable boundary behavior for robust evaluation: 

(i) \textbf{Ideal Utilization ($NCU=1$):} Occurs when the posterior probability equals 1, meaning the retrieved context provides total certainty for the target sequence. 

(ii) \textbf{Baseline Nullity ($NCU=0$):} Occurs when the posterior confidence is less than or equal to the prior, implying the context provides zero or negative information gain. 

(iii) \textbf{Sensitivity:} The derivative of NCU with respect to the posterior log-probability is constant, indicating that the metric scales linearly with the model’s confidence gain until the entropy is fully resolved.

\section{Experimental Setup}
\label{sec:experimental_setup}

To effectively isolate active contextual processing from parametric memory, we engineered a multi-domain evaluation framework anchored in continuous, length-normalized token-level log-probabilities. 

\subsection{Datasets and Task Selection}
\label{subsec:datasets}

We evaluated our framework across three established Question Answering (QA) corpora \cite{jiang2023active, ram2023incontext} to ensure a comprehensive assessment of extractive capabilities. Specifically, we utilized Natural Questions (NQ-Open) to evaluate real-world, single-hop entity extraction, alongside the \textit{rc.wikipedia} split of TriviaQA to assess entity-centric factoid extraction from structured contexts. We also incorporated the \textit{distractor} split of HotpotQA, a dataset typically requiring reasoning across interconnected documents; however, due to the strict token generation limits imposed in our study to mathematically isolate pure entity extraction, multi-hop synthesis was inherently constrained. 

To isolate the interaction between retrieved context and parametric memory, these evaluations were deliberately restricted to monolingual English corpora. Expanding to multilingual or low-resource languages introduces significant analytical noise driven by differing tokenization efficiencies and inherently weaker parametric priors. By strictly evaluating in English, where LLM parametric priors are overwhelmingly dominant, we subjected the RAG extraction process to the most stringent epistemic stress test possible. Furthermore, we employed stratified subsampling across these corpora to ensure statistical parity across reasoning typologies.

\subsection{Model Selection and Paradigmatic Evaluation}
\label{subsec:models}
To test the universality of scaling laws in extraction tasks, we selected architectures that allow both strict intra-family scaling comparisons and cross-paradigm benchmarking:
\begin{enumerate}
    \item \textbf{SLM Baseline:} \texttt{Qwen2.5-1.5B-Instruct} \cite{qwen2024qwen25}. Functions as an efficient, edge-deployable dense model.
    \item \textbf{Mid-Weight Model:} \texttt{Qwen2.5-7B-Instruct}. Enables measurement of the marginal utility of a $\sim 5\times$ parameter increase within the identical architecture family.
    \item \textbf{Massive Open-Weights Baseline:} \texttt{Qwen2.5-72B-Instruct}. Serves as our primary scaling upper-bound, isolating the effects of pure parameter scale while holding the pre-training philosophy and dense architecture relatively constant.
    \item \textbf{Proprietary Commercial Archetype:} \texttt{gpt-4o-mini}. Included not for direct 1:1 architectural comparison, but as an archetypal representative of the closed-API paradigm: high-throughput, likely utilizing Mixture of Experts (MoE), and heavily subjected to proprietary safety alignments (e.g., RLHF).
\end{enumerate}

Local inferences were executed on NVIDIA H100 hardware. The SLMs were run in \texttt{float16} precision, while the 72B model utilized quantized precision to accommodate memory constraints without meaningful performance degradation. Generation was strictly bounded to 5 tokens using greedy decoding ($T=0.0$). A uniform zero-shot directive prompt was applied across all architectures to prevent model-specific prompt engineering from confounding the baseline epistemic behavior.

\subsection{Context Perturbation Engine}
\label{subsec:perturbation}

\begin{figure}[htb!]
    \centering
    \resizebox{0.9\textwidth}{!}{
    \begin{tikzpicture}[
        font=\sffamily\small,
        box/.style={align=center, inner sep=2mm, rounded corners=4pt, minimum width=2.5cm},
        input/.style={box, fill=cyan!15, draw=cyan!50, line width=1pt},
        process/.style={box, fill=teal!10, draw=teal!50, line width=1.5pt, dashed},
        output/.style={box, fill=purple!10, draw=purple!50, line width=1pt},
        arrow/.style={-{Stealth[scale=1.0]}, thick, draw=black!60}
    ]

    \node (raw) [input] {\textbf{Original Tuple} \\ $(Q, Y, C_{oracle})$};
    \node (engine) [process, below=0.7cm of raw] {\textbf{Perturbation Engine} \\ (NER + Synthetic Noise)};
    
    \node (oracle) [output, below=0.9cm of engine, xshift=-1.8cm] {\textit{Oracle} \\ ($Q + C_{oracle}$)};
    \node (noise) [output, below=0.9cm of engine, xshift=1.8cm] {\textit{Noise} \\ ($Q + C_{distractor}$)};
    \node (zero) [output, left=0.4cm of oracle] {\textit{Zero-Shot} \\ ($Q$)};
    \node (adv) [output, right=0.4cm of noise] {\textit{Adversarial} \\ ($Q + C_{conflict}$)};

    \draw [arrow] (raw) -- (engine);
    \draw [arrow] (engine.south) -- (zero.north);
    \draw [arrow] (engine.south) -- (oracle.north);
    \draw [arrow] (engine.south) -- (noise.north);
    \draw [arrow] (engine.south) -- (adv.north);

    \end{tikzpicture}
    }
    \caption{\textbf{Context Perturbation Engine Architecture.} The transformation of a single evaluation tuple into four controlled experimental conditions, aiming to isolate parametric memory from active context utilization.}
    \label{fig:perturbation_engine}
\end{figure}
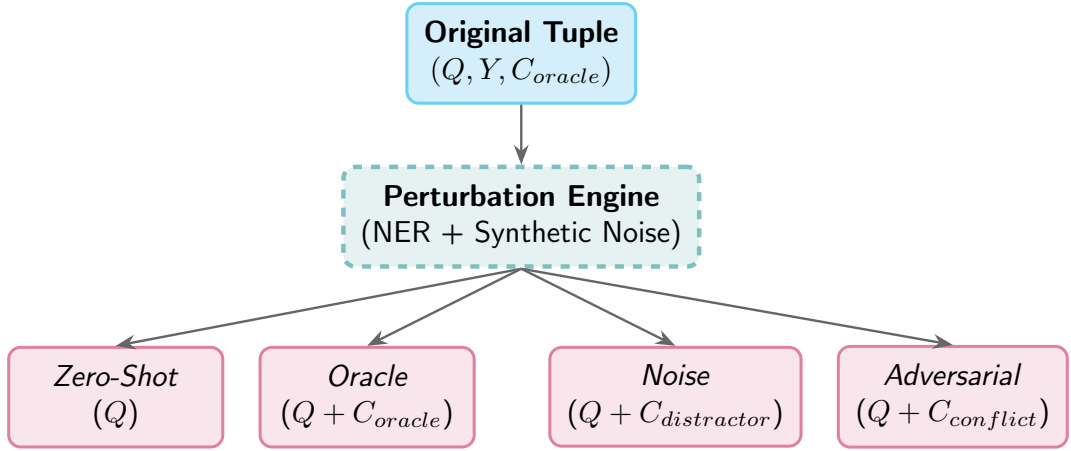

As illustrated in Figure \ref{fig:perturbation_engine}, for each question-answer pair $(q, y)$, we synthesized four distinct conditions:
\begin{itemize}
    \item \textbf{Zero-Shot ($c_{zero}$):} The model is prompted solely with the query, establishing baseline predictive uncertainty.
    \item \textbf{Oracle ($c_{oracle}$):} The model is provided the ground-truth context containing the exact answer.
    \item \textbf{Adversarial Conflict ($c_{adv}$):} Utilizing a Named Entity Recognition (NER) pipeline, we systematically replaced the true answer in the oracle context with a fabricated entity. This quantifies \textit{Prior Dominance}.
    \item \textbf{Semantic Noise ($c_{noise}$):} The oracle context is embedded centrally within semantically irrelevant synthetic text, acting as a control for context-length degradation.
\end{itemize}

\subsection{Evaluation Metrics: Normalization Strategy}
\label{subsec:metrics}
To mitigate length bias, sequence confidence is calculated as the arithmetic mean of the target token log-probabilities, ensuring equitable evaluation across models with distinct tokenization vocabularies. For discrete accuracy metrics, a generation is classified as a successful extraction strictly if its length-normalized probability satisfies $P > 0.05$. Given standard LLM vocabularies exceeding 100,000 tokens (where uniform random selection yields $P \approx 10^{-5}$), this conservative threshold isolates deterministic extraction by ensuring the model's predictive confidence operates magnitudes above stochastic noise.

\section{Results and Analysis}
\label{sec:results}

\begin{figure}[htb!]
    \centering
    \includegraphics[width=\textwidth]{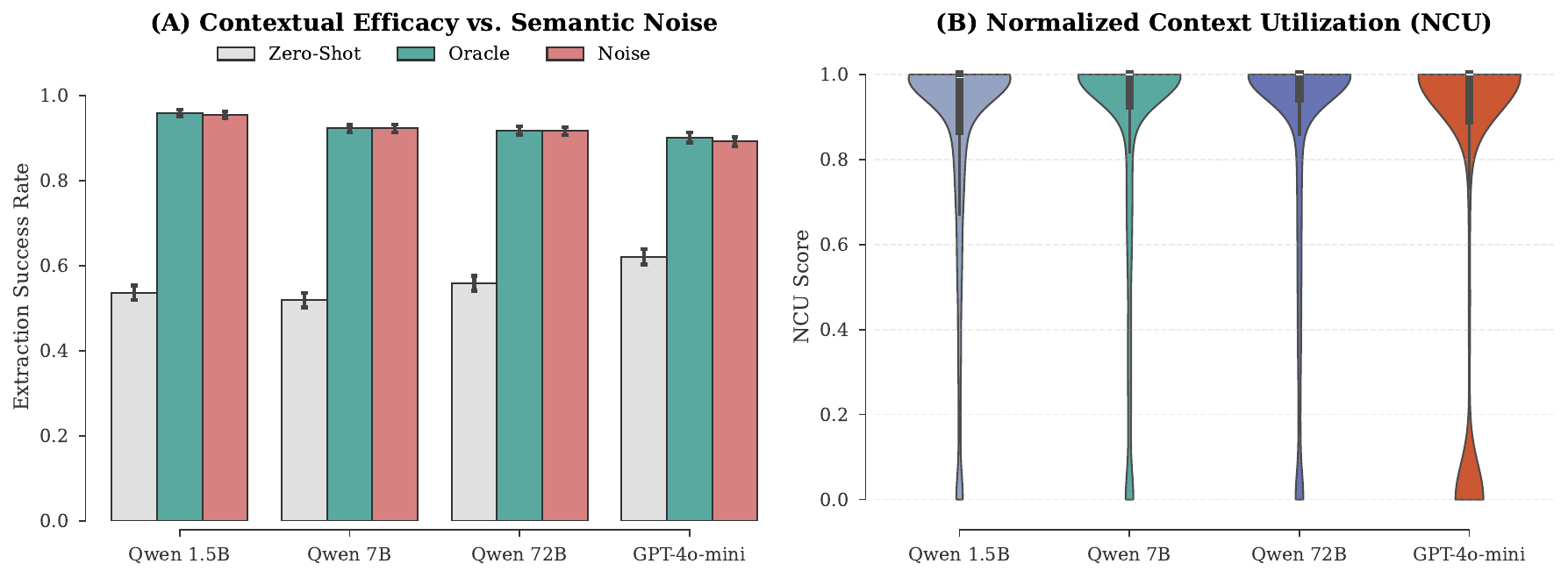}
    \caption{\textbf{Contextual Efficacy and NCU Distribution.} (A) The discrete success rates across Zero-Shot, Oracle, and Noise conditions. SLMs exhibit highly competitive context-grounded extraction. (B) The distribution of the bounded NCU scores, demonstrating the stable and high information gain achieved by the 1.5B and 7B architectures.}
    \label{fig:utilization}
\end{figure}

\begin{figure}[htb!]
    \centering
    \includegraphics[width=\textwidth]{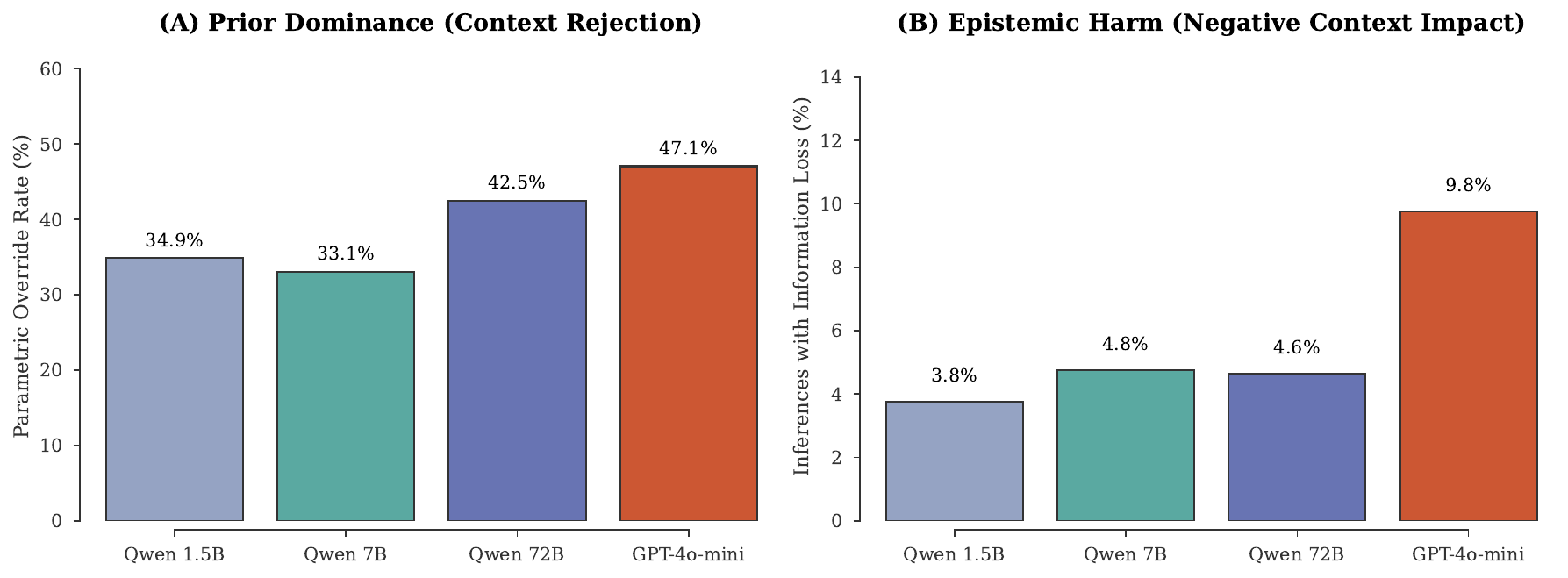}
    \caption{\textbf{Prior Dominance and Information Loss.} (A) Prior Dominance: The percentage of inferences where the model overrode explicit adversarial context in favor of its prior parametric knowledge. (B) Negative Transfer: The frequency of instances where confusing context actively degraded the model's predictive confidence below its zero-shot baseline.}
    \label{fig:stubbornness}
\end{figure}

Our empirical evaluation confirms a structural divergence between parametric scale and contextual processing efficacy. The results, synthesized in Table \ref{tab:summary_results} and Figures \ref{fig:utilization} and \ref{fig:stubbornness}, address our core research questions by revealing diminishing returns in traditional scaling paradigms for pure extraction tasks.

\begin{table}[htbp]
    \centering
    \caption{\textbf{Empirical Summary of Contextual Extraction.} Results derived from the balanced evaluation dataset (1,000 samples per corpus). Accuracy metrics represent discrete success rates ($P>0.05$). To rigorously handle outliers, the bounded NCU is clamped within $[0,1]$ \textit{per-inference} prior to averaging, whereas the Raw NCU average remains unclipped to expose the magnitude of negative transfer.}
    \vspace{2mm}
    \resizebox{\textwidth}{!}{
    \begin{tabular}{@{}lccccccc@{}}
        \toprule
        \textbf{Model} & \textbf{Zero-Shot} & \textbf{Oracle} & \textbf{Noise} & \textbf{Conflict} & \textbf{NCU} & \textbf{Raw NCU} & \textbf{Latency} \\ 
        & $(acc\_zero)$ & $(acc\_oracle)$ & $(acc\_noise)$ & $(acc\_conflict)$ & & & (s/q) \\ \midrule
        Qwen 1.5B & 53.6\% & \textbf{95.9\%} & \textbf{95.5\%} & 34.9\% & 0.864 & 0.620 & \textbf{0.041} \\
        Qwen 7B & 51.9\% & 92.3\% & 92.3\% & \textbf{33.1\%} & 0.864 & -74.128 & 0.093 \\
        Qwen 72B & 55.9\% & 91.8\% & 91.7\% & 42.5\% & \textbf{0.868} & -20.882 & 0.389 \\
        GPT-4o-mini & \textbf{62.0\%} & 90.1\% & 89.2\% & 47.1\% & 0.777 & -7779.254 & 2.982 \\
        \bottomrule
    \end{tabular}
    }
    \label{tab:summary_results}
\end{table}

\subsection{The Utilization Gap and Scaling Returns (RQ1)}
Our primary objective was to ascertain whether scaling up model parameters intrinsically amplifies contextual extraction. The empirical data indicates severe diminishing returns for factual grounding. While the commercial baseline (\texttt{gpt-4o-mini}) achieved the highest zero-shot accuracy (62.0\%), indicating superior parametric memorization, it exhibited the lowest oracle extraction accuracy (90.1\%) and the lowest bounded NCU score (0.777). 

In contrast, the highly efficient \texttt{Qwen-1.5B} model acted as an optimal contextual processor. It achieved a peak oracle accuracy of 95.9\% and maintained an NCU of 0.864, effectively matching the continuous extraction capabilities of its 72B counterpart (NCU 0.868) while operating at a fraction of the computational cost (0.041s vs. 0.389s latency per query). This confirms that for strict extraction tasks, the relatively unconstrained nature of SLMs provides a structural efficiency advantage over heavily parameterized models.

\subsection{Prior Dominance and Adherence (RQ2)}
We investigated whether the depth of a model's parametric memory, or its proprietary alignment, induces a measurable resistance to integrating adversarial context. This phenomenon, measured as \textit{Prior Dominance}, is illustrated in Figure \ref{fig:stubbornness}A.

When the oracle context was perturbed to state a fabricated entity, the \texttt{Qwen-7B} and \texttt{1.5B} models exhibited high contextual adherence, surviving the conflict by adhering strictly to the provided text in roughly 65-67\% of cases. Conversely, the commercial baseline rejected the adversarial context and defaulted to its parametric prior in 47.1\% of the conflicts, despite explicit prompt instructions to synthesize answers strictly from the retrieved text. The open-weights \texttt{Qwen-72B} exhibited a 42.5\% parametric override rate. This progression implies that while prior dominance is heavily influenced by parameter scale (as seen in the jump from 7B to 72B), it may be further exacerbated by safety-oriented alignments present in proprietary APIs.

\subsection{Noise Resilience and Negative Transfer (RQ3)}
To ensure our NCU measurements were not artificially deflated by context length degradation, we evaluated the models' resilience to semantic noise. While discrete accuracy remained relatively stable across architectures when noise was introduced (Table \ref{tab:summary_results}), analyzing the unbounded \textit{Raw NCU} revealed critical vulnerabilities beneath the surface.

Mathematically, a negative Raw NCU occurs when the posterior entropy given the retrieved context strictly exceeds the prior zero-shot entropy ($S_{post} > S_{prior}$). In other words, rather than resolving uncertainty, the external context acted as a destructive interference that actively degraded the model's confidence below its initial ignorance. From a Bayesian perspective, this collapse in confidence is an expected "surprise" when a highly confident prior is contradicted. However, operationally within a RAG pipeline, this constitutes a systemic vulnerability. The extreme negative magnitude observed in the raw average of the commercial baseline (-7779.254) is a mathematical artifact of its high initial confidence: dividing a negative information gain by a near-zero prior entropy ($S_{prior} \approx 0$) yields an asymptotic penalty.

As shown in Figure \ref{fig:stubbornness}B, we quantified this \textit{Negative Transfer}. The proprietary baseline suffered mathematical information loss in 9.8\% of inferences, a rate nearly triple that of the \texttt{Qwen-1.5B} model (3.8\%). This indicates that when high-capacity LLMs encounter conflicting or confusing out-of-distribution knowledge, their internal probability distributions can systemically collapse, rendering them highly unpredictable in strict processing workflows compared to SLM counterparts.

\section{Discussion}
\label{sec:discussion}

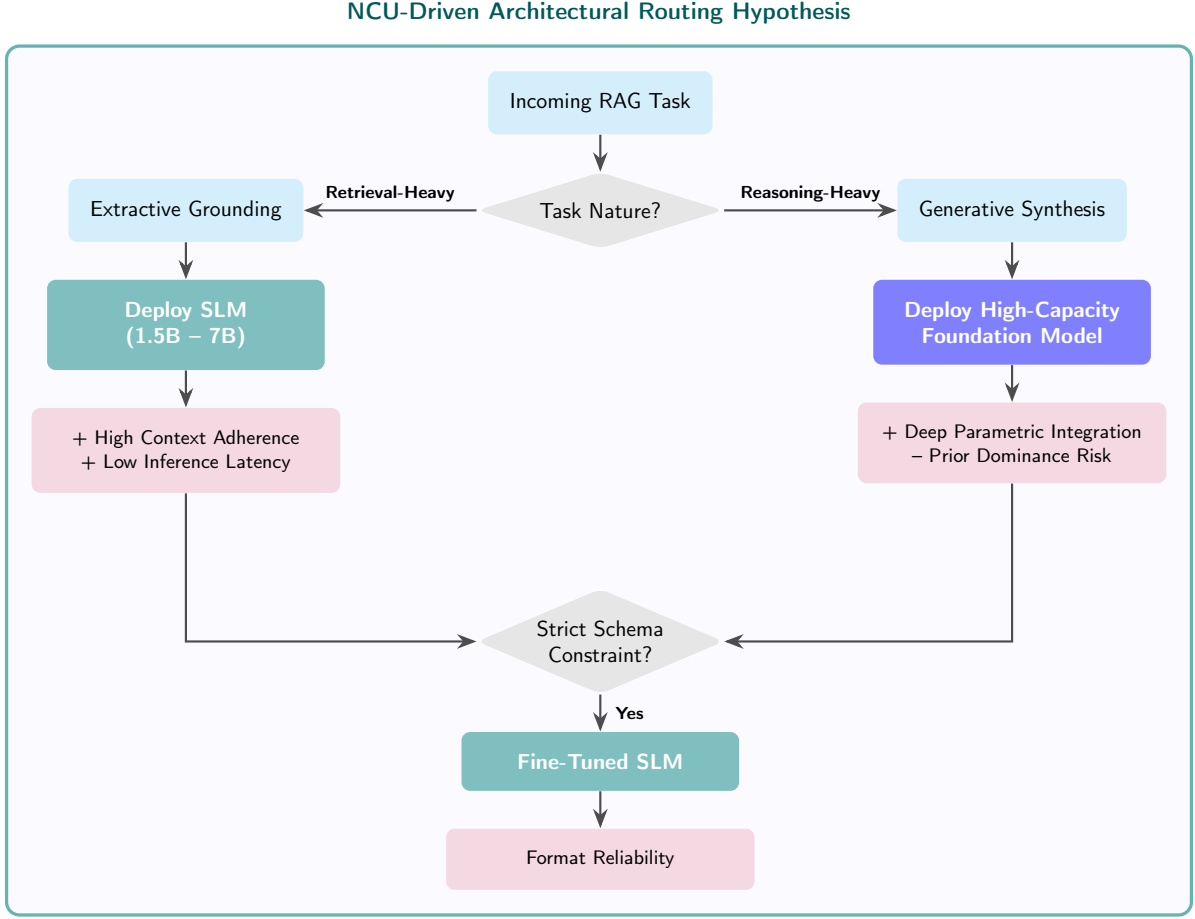
\begin{figure}[htb!]
    \centering
    \resizebox{\textwidth}{!}{
    \begin{tikzpicture}[
        font=\sffamily\small,
        base/.style={draw=none, align=center, inner sep=3.5mm, rounded corners=4pt},
        decision/.style={base, diamond, fill=gray!20, aspect=2, inner sep=0pt, minimum width=4cm},
        model_slm/.style={base, fill=teal!50, text=white, minimum width=4.5cm, font=\sffamily\small\bfseries},
        model_massive/.style={base, fill=blue!50, text=white, minimum width=4.5cm, font=\sffamily\small\bfseries},
        data/.style={base, fill=cyan!15, text=black},
        outcome/.style={base, fill=purple!15, text=black, minimum width=5cm, font=\sffamily\footnotesize},
        arrow/.style={-{Stealth[scale=1.2]}, line width=1pt, draw=black!70},
    ]

    \node (start) [data] {Incoming RAG Task};
    \node (dec1) [decision, below=0.6cm of start] {Task Nature?};
    
    \node (path_ext) [data, left=2.8cm of dec1] {Extractive Grounding};
    \node (slm) [model_slm, below=0.6cm of path_ext] {Deploy SLM \\ (1.5B -- 7B)};
    \node (out_slm) [outcome, below=0.6cm of slm] {+ High Context Adherence \\ + Low Inference Latency};

    \node (path_reason) [data, right=2.8cm of dec1] {Generative Synthesis};
    \node (massive) [model_massive, below=0.6cm of path_reason] {Deploy High-Capacity \\ Foundation Model};
    \node (out_massive) [outcome, below=0.6cm of massive] {+ Deep Parametric Integration \\ -- Prior Dominance Risk};

    \node (dec2) [decision, below=5.5cm of dec1] {Strict Schema \\ Constraint?};
    
    \node (ft_slm) [model_slm, below=0.6cm of dec2] {Fine-Tuned SLM};
    \node (out_ft) [outcome, below=0.6cm of ft_slm] {Format Reliability};

    \draw [arrow] (start.south) -- (dec1.north);
    
    \draw [arrow] (dec1.west) -- node[above, font=\sffamily\scriptsize\bfseries] {Retrieval-Heavy} (path_ext.east);
    \draw [arrow] (dec1.east) -- node[above, font=\sffamily\scriptsize\bfseries] {Reasoning-Heavy} (path_reason.west);

    \draw [arrow] (path_ext.south) -- (slm.north);
    \draw [arrow] (slm.south) -- (out_slm.north);

    \draw [arrow] (path_reason.south) -- (massive.north);
    \draw [arrow] (massive.south) -- (out_massive.north);

    \draw [arrow] (out_slm.south) |- (dec2.west);
    \draw [arrow] (out_massive.south) |- (dec2.east);
    
    \draw [arrow] (dec2.south) -- node[right, font=\sffamily\scriptsize\bfseries, xshift=0.1cm] {Yes} (ft_slm.north);
    \draw [arrow] (ft_slm.south) -- (out_ft.north);

    \begin{scope}[on background layer]
        \node (box) [draw=teal!60, line width=1.5pt, rounded corners=6pt, fill=blue!2, inner sep=4mm, fit=(start) (path_ext) (out_slm) (path_reason) (out_massive) (out_ft)] {};
        \node [above=2mm, font=\sffamily\bfseries\color{teal!70!black}] at (box.north) {NCU-Driven Architectural Routing Hypothesis};
    \end{scope}

    \end{tikzpicture}
    }
    \caption{\textbf{Conceptual RAG Operational Routing.} Based on NCU evaluations, extractive tasks may be efficiently directed to SLMs to maximize context adherence and reduce latency. High-capacity models remain optimal for complex synthesis, provided the risk of parametric overwriting is managed.}
    \label{fig:ncu_routing_framework}
\end{figure}

The deployment of LLMs has historically relied on the heuristic that parametric scaling uniformly enhances all downstream capabilities. However, our evaluation using the NCU metric highlights a notable structural divergence: pure contextual extraction appears to operate on fundamentally different epistemic mechanisms from generalized reasoning. 

This divergence manifests through the observed risk of \textit{Negative Transfer} and Prior Dominance. We introduce the \textbf{"Alignment Tax Hypothesis"} to frame this behavior: the evaluated commercial baseline exhibited a measurable tendency to default to its pre-training distributions when external evidence contradicted its high-confidence priors. In these instances, the model did not merely fail to extract the new information; its predictive probability distribution exhibited severe entropy degradation, leading to a negative transfer of information (a 9.8\% negative transfer rate). While the extreme mathematical magnitude of the raw NCU penalty (-7779.25) is primarily an artifact of dividing by a near-zero initial entropy, the systemic confidence collapse remains a tangible vulnerability. It suggests a potential "tax" where heavy safety alignments or dense parametric memorization actively resist out-of-distribution updates.

Conversely, SLMs functioned more closely as unconstrained processors. They exhibited higher context adherence, robust noise resilience, and suffered less than half the negative transfer rate (3.8\%). Achieving statistical parity in bounded NCU scoring (0.864 for 1.5B vs. 0.868 for 72B) alongside significantly lower inference latency (0.041s vs. 0.389s) suggests severe diminishing returns for parameter scaling in strict extraction tasks. These findings theoretically advocate for decoupled, modular RAG architectures where task routing is determined by epistemic requirements rather than a default reliance on high-capacity architectures (Figure \ref{fig:ncu_routing_framework}).

\section{Limitations and Future Work}
\label{sec:limitations}

While the NCU framework rigorously quantifies relative context utilization, our experimental design has distinct analytical boundaries. First, the strict token generation limit ($T \le 5$), necessary to mathematically isolate pure extraction probabilities, restricts our claims strictly to factual extraction. It precludes the evaluation of multi-hop synthesis or Chain-of-Thought (CoT) paradigms, where high-capacity models typically demonstrate emergent reasoning capabilities.

Second, our analysis of epistemic stubbornness relies on correlational observations rather than strict causal ablations. Comparing open-weight dense models directly to proprietary APIs introduces confounding variables such as Mixture of Experts (MoE) routing, undisclosed pre-training data mixtures, and unknown tokenization granularities. The opacity of the commercial baseline means that the exact mechanisms driving its negative transfer (e.g., specific RLHF penalties, constitutional AI constraints, or attention dilution) remain hypothesized under our "Alignment Tax" framework.

Finally, the adversarial conflicts generated via the Faker library are inherently synthetic. It is plausible that highly parameterized models possess latent anomaly detection capabilities, implicitly recognizing and rejecting these semantic forgeries rather than failing to comprehend the instruction.

\textbf{Future Work:} Subsequent research must conduct strict ablation studies on open-weights models (e.g., evaluating a Base model vs. its RLHF-Instruct counterpart) to definitively isolate the causal impact of safety alignment on Prior Dominance. Furthermore, expanding the evaluation across multiple commercial endpoints (e.g., Claude, Gemini) and diverse linguistic corpora will verify the universality of these epistemic vulnerabilities. Extending the length-normalized NCU metric to evaluate character-level cross-entropy and free-form generative reasoning warrants immediate exploration.

\section{Conclusion}
\label{sec:conclusion}

This study introduces the Normalized Context Utilization (NCU) metric to address the epistemic blindness inherent in discrete RAG evaluations. By analyzing continuous, length-normalized token probabilities across perturbed conditions, we observed significant diminishing returns associated with traditional parameter scaling for pure extraction tasks. The evaluated commercial model frequently exhibited Prior Dominance, actively losing predictive confidence when exposed to adversarial external evidence.

Conversely, SLMs in the 1.5B to 7B parameter regime demonstrated highly competitive contextual utilization and superior factual adherence. Coupling these epistemic characteristics with substantial improvements in inference latency suggests that deploying high-capacity, highly-aligned models for strict factual grounding may be computationally sub-optimal. The empirical insights derived from the NCU framework suggest that in the design of robust RAG pipelines, parametric scale must be carefully balanced with the need for epistemic flexibility.

\section*{Code Availability}
To facilitate reproducibility and further research, the code, experimental framework, and datasets associated with this study are publicly available at \url{https://github.com/BarakOr1/Quantifying-Prior-Dominance-in-RAG-Systems}.

\section*{Acknowledgments}
This research was independently conceived, conducted, and funded by ArtificialGate Ltd. All computational resources, proprietary data processing, and research hours were provided exclusively by the author's private firm. No institutional funding, facilities, or support from the author's academic affiliations were utilized in the development or execution of this study.

\bibliographystyle{unsrt} 
\bibliography{references}
\end{document}